\documentclass[10pt,english]{article}
\usepackage[T1]{fontenc}
\usepackage{graphicx}
\usepackage{setspace}

\makeatletter

 \usepackage{verbatim}

\date{}
\usepackage{times}
\usepackage{llncsdoc}

\usepackage{fancyhdr}
\makeatother
\begin{document}

\newcommand{\bfrac}[2]{\frac{#1 }{#2 }}

\newcommand{\cA}{{\mathcal{A}}}

\newcommand{\cD}{\mathcal{D}}

\newcommand{\cH}{\mathcal{H}}

\newcommand{\cS}{\mathcal{S}}

\newcommand{\cT}{\mathcal{T}}

\newcommand{\cX}{\mathcal{X}}

\newcommand{\dsb}[1]{{\left[\!\left[#1\right]\!\right]}}

\newcommand{\eps}{\varepsilon}

\newcommand{\es}{\emptyset}

\newcommand{\Loss}{\textrm{Loss}}

\newcommand{\mydiff}[1]{\bfrac{\partial}{\partial#1}}

\newcommand{\R}{\mathbb{R}}

\newcommand{\Sign}{\textrm{Sgn}}

\newcommand{\nth}{^{\textrm{th}}}

\newcommand{\vst}{\vspace{3mm}}

\newcommand{\mycomment}[1]{}

\title{A Theory of Probabilistic Boosting,\\ Decision Trees and Matryoshki
\vspace{-3mm}}

\author{Etienne Grossmann\\
visiting at Université de Montréal\\
\texttt{\small <etienne@isr.ist.utl.pt>}\vspace{-5mm}}

\maketitle
\begin{abstract}
We present a theory of boosting probabilistic classifiers. We place
ourselves in the situation of a user who only provides a stopping
parameter and a probabilistic weak learner/classifier and compare
three types of boosting algorithms: probabilistic Adaboost, decision
tree, and tree of trees of ... of trees, which we call \emph{matryoshka}.
{}``Nested tree,'' {}``embedded tree'' and {}``recursive tree''
are also appropriate names for this algorithm, which is one of our
contributions. Our other contribution is the theoretical analysis
of the algorithms, in which we give training error bounds. This analysis
suggests that the matryoshka leverages probabilistic weak classifiers
more efficiently than simple decision trees.\vspace{-1mm}
\end{abstract}
\thispagestyle{fancy}

\section{Introduction\vspace{-1mm}}

Ensembles of classifiers are a popular way to build a strong classifier
by leveraging simple decision rules -weak classifiers. Many ensemble
architectures have been proposed, such as neural networks, decision
trees, Adaboost~\cite{SchapireSinger99ML}, bagged classifiers~\cite{Breiman96ML},
random forests~\cite{Breiman01ML}, trees holding a boosted classifier
at each node~\cite{Tu05ICCV}, boosted decision trees... One drawback
of ensemble methods is that they are often dispendious about the computational
cost of the resulting classifier. For example, Adaboost~\cite{SchapireSinger99ML},
bagging~\cite{Breiman96ML}, random forests~\cite{Breiman01ML}
all multiply the runtime complexity, by a factor approximately proportional
to the training time.This is not acceptable in applications involving
large amounts of data and requiring low-complexity method, such as
 video analysis and data mining.

Many approaches have been proposed to deal with such situations. The
cascade architecture, i.e. a degenerate decision tree, has become
very popular~\cite{ViolaJones01ICCV} and has been intensely studies~\cite{McCaneNovins03ICV,Luo05CVPR}.
However, cascades are mostly appropriate to detect rare exemplars
of interest amongst a huge majority of uninteresting ones. 

Decision trees, on the other hand, are better adapted to the case
of balanced target classes. This advantage comes from their greater
facility to decompose the input space into more manageable and useful
subsets. In addition, their run-time complexity is approximately proportional
to the logarithm of the training time. Counterbalancing these advantages,
is the fact that decision trees tend to overfit the training data.

There exist many proposed methods to improve overfitting, for example
pruning and smoothing, but the main recognized cause remains: data
elements are passed to one only of the descendant of each node, whether
during training, or at run-time. At run-time, one proposed solution
is to pass examples along more than one child node~\cite{ChangPavlidis77TSMC}.
During training, it has been proposed~\cite{Tu05ICCV} to pass down
to all descendants the exemplars that lay within a fixed distance
of the separating surface.

Our approach to avoid the hard split at each node is to consider that
the examples have a certain probability -not necessarily always 0
or 1- of being passed to any descendant of the tree. That is, we study
probabilistic decision trees~\cite{Quinlan90MLAIA}, but pursue a
different analysis from these last works. First, we show that probabilistic
decision trees are eminently tractable within the framework of boosting%
\footnote{The analogy between deterministic decision trees and boosting has
been studied in~\cite{KearnsMansour99JCSS}.%
}.

We bound the expected misclassification error as a function of the
number of nodes in the tree, in Section~\ref{sec:boosting-DT-model}.
This bound is very high when the probabilistic weak classifiers are
very weak. Moreover, we present arguments that suggest that any bound
using the same probabilistic weak learner hypothesis will necessarily
be high.

However, we also note that the bound achieved with stronger weak classifiers
is much better. In an attempt to strengthen our weak classifiers,
we explore the possibility of assembling decision trees consisting
of decision trees, the inner and the outer trees being built by the
same algorithm. This is not the first time this idea is suggested,
but we believe we are the first to show the theoretical benefits of
doing so.

\begin{figure}
\begin{center}\hfill{}\includegraphics[%
  width=0.5\textwidth,
  height=2.8cm]{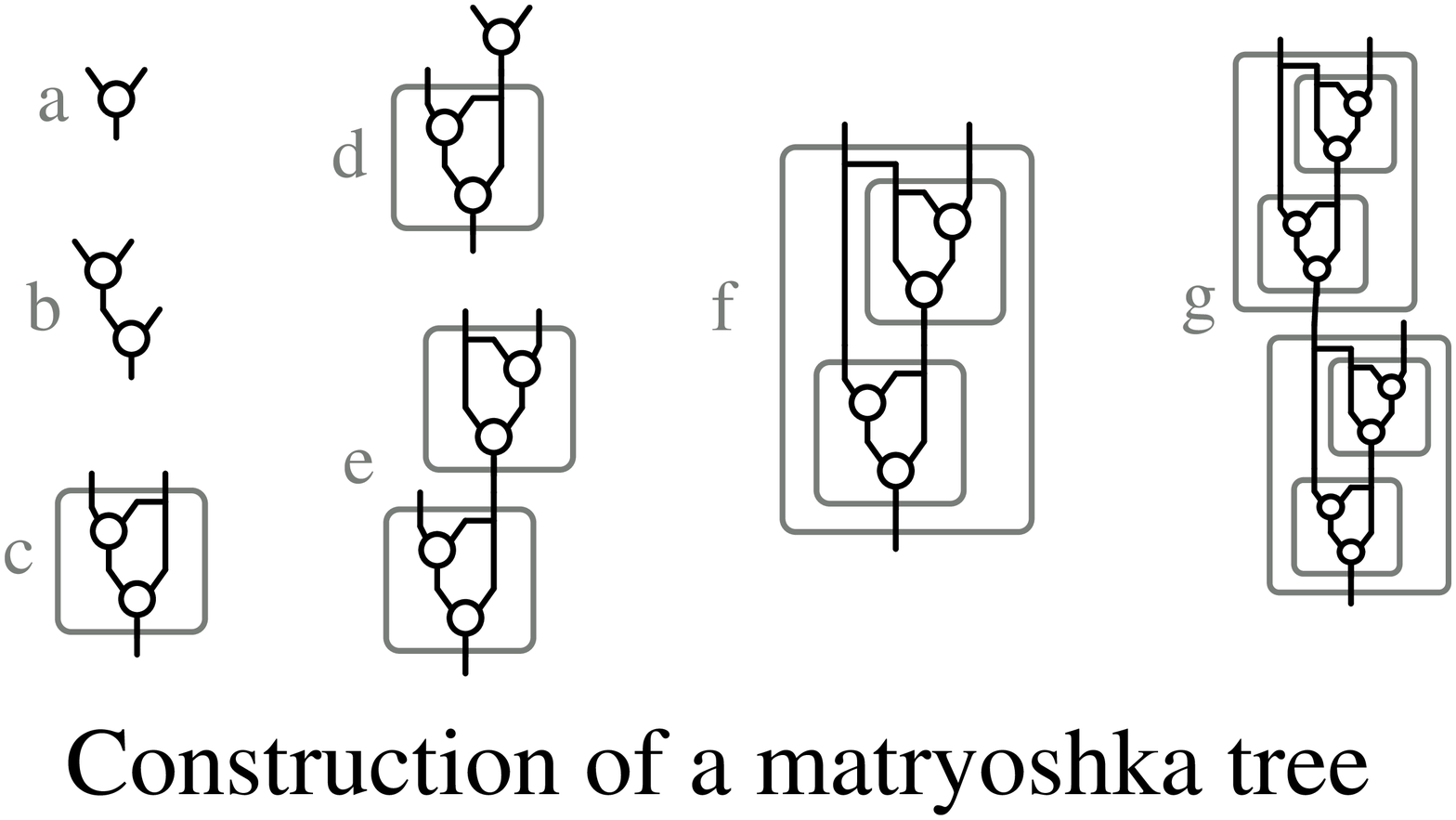}\hfill{}\includegraphics[%
  width=0.19\textwidth,
  keepaspectratio]{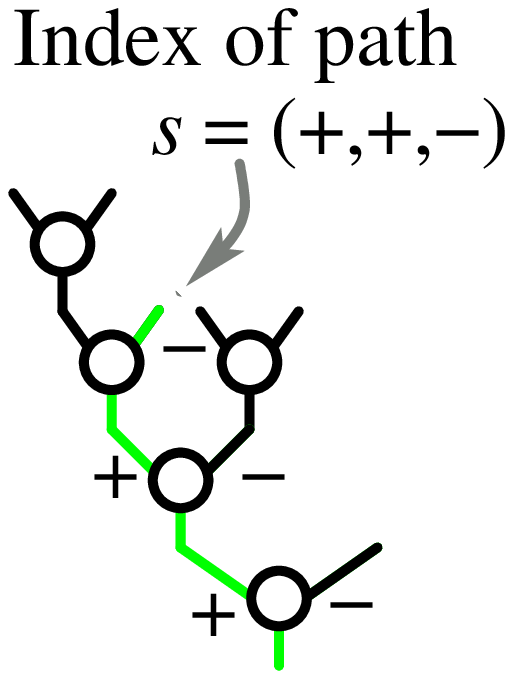}\hfill{}\vspace{-3mm}\end{center}

\caption{Left: Construction of a matryoshka decision trees. Here, each tree
has just two nodes, but other numbers are possible\label{cap:Nested-decision-tree}.
Right: notation use to specify a path in a decision tree\label{fig:decisionTreeIndex}.\vspace{-2mm}}
\end{figure}

Continuing on the idea of embedding (or nesting) decision trees one
into another, we propose to assemble trees of trees of ... of trees
of probabilistic weak classifiers. This is similar to matryoshka dolls,
with the difference that each tree contains more than one tree, rather
than a single other doll. Figure~\ref{cap:Nested-decision-tree},
left, illustrates this concept. Our main contribution (Section~\ref{sub:Bound-Matryoshka})
is to prove a greatly improved bound, reached by trees with exactly
two nodes, each node being a tree with two nodes, and so on until
the last nesting level, which holds two probabilistic weak classifiers.

Another merit of our study is that it proposes a methodology that
is essentially parameterless. The user only needs to provide a probabilistic
weak learner and a stopping criterion, such as the number of nodes
or the error on the training dataset. If a stopwatch%
\footnote{This metaphor is to say that, if the time complexity of the learner
and classifier are known or measurable, then this information can
be used to greedily reduce the training error.%
}, is available during training, then we propose ways of using it.
The freedom of parameters results partly from applying a principle
of greedy error minimization.

Before presenting our study on decision trees, we define, in Section~\ref{sec:Probabilistic-weak-learner},
the probabilistic weak learners that are the basis of this work. We
then present, in Section~\ref{sec:Adaboost}, the probabilistic equivalent
of Adaboost that will serve as reference for the rest of the article.
After presenting our main theory in Sections~\ref{sec:boosting-DT-model}
and \ref{sec:Matryoshka}, we discuss further the findings of this
study and open directions for future research.

\section{Probabilistic weak learner\label{sec:Probabilistic-weak-learner}\vspace{-1mm}}

We consider learning algorithms $\cA$ that, given a training dataset
$\cS=\left\{ \left(X_{1},y_{1},D(1)\right),\right.$ $\left(X_{2},y_{2},D(2)\right),\ldots,$
$\left.\left(X_{N},y_{N},D(N)\right)\right\} ,$ with $D\left(1\right)+\ldots+D\left(N\right)=1$,
return a probabilistic classifier, or oracle, written $h\left(X\right)$.
For any input $X$, $h\left(X\right)$ is identified with a Bernoulli
random variable with parameter $q\left(+,X\right)$.

\begin{description}
\item [Definition:]We say that $\cA$ is a \emph{probabilistic weak learner},
if there exists a constant $0<\eps\leq\frac{1}{2}$ such that, for
any dataset $\cS$, the expected error of $h\left(X\right)$, is smaller
than $\frac{1}{2}-\eps$; that is, one has:\[
\sum_{n=1}^{N}D\left(n\right)q\left(-y_{n},X_{n}\right)\leq\frac{1}{2}-\eps,\]
where $q\left(-y_{n},X_{n}\right)$ is the probability that $h\left(X_{n}\right)$
takes the value $-y_{n}$, i.e. that the classifier is wrong.
\end{description}
The constant $\eps$, called the \emph{advantage} or \emph{edge} is
unknown and does not need to be known. The probability $q\left(y_{n},X_{n}\right)$,
also unknown, will be needed. We estimate it by calling repeatedly
the weak classifier and calculating the maximum likelihood (ML) or
maximum a-posteriori (MAP) estimates: If $h_{1,n},\ldots,h_{R,n}$
are the values returned by $R$ invocations (observations) of $h\left(X_{n}\right)$,
then the ML estimate is$\tilde{q}\left(y,X_{n}\right)=\left|\left\{ n\,\mid\, h_{r,n}=y\right\} \right|/R$,
where $\left|.\right|$ is the set cardinal. Assuming that $q\left(y,X_{n}\right)$
is uniformly distributed in $\left[0,1\right]$%
\footnote{This prior is pessimistic, since $q\left(y_{n},X_{n}\right)$ has
(unknown) expectation smaller than $\frac{1}{2}$, but the edge $\eps$
being unknown, using another prior would not be less hazardous.%
}, the MAP is $\hat{q}\left(y,X_{n}\right)=$$\frac{1}{R+2}\left(1+\left|\left\{ n\,\mid\, h_{r,n}=y\right\} \right|\right)$.

\section{Adaboost for probabilistic weak learners\label{sec:Adaboost}\vspace{-1mm}}

We now adapt Adaboost~\cite{SchapireSinger99ML} to probabilistic
-rather than deterministic- weak learners. \mycomment{ Our assumptions
should not be confused with that of stochastic gradient boosting~\cite{Friedman02CSDA}.}
Like the original Adaboost, we consider classifiers of the form

\[
H\left(X\right)=\sum_{t=1}^{T}\alpha_{t,X}h_{t}\left(X\right),\]
but here, $h_{t}\left(X\right)$ is a Bernoulli random variable (or
randomized classifier or oracle), so that $H\left(X\right)$ is itself
a random variable. Like in Adaboost, we consider domain-partitioned
weights (see \cite[ Sec. 4.1]{SchapireSinger99ML}): we have constants
$\alpha_{t,+}$ and $\alpha_{t,-}$ such that $\alpha_{t,X}=\alpha_{t,+}$
if $h_{t}\left(X\right)$ is observed to be $+1$, and that $\alpha_{t,X}=\alpha_{t,-}$
otherwise. 

We proceed as in Adaboost, increasing the number $T$ of weak classifier,
and not changing a weak classifier once it has been trained. Each
random classifier $h_{t}\left(X\right)$ is obtained by running the
weak learner on the training data set $\left(X_{1},y_{1}\right),$$\ldots,$$\left(X_{N},y_{N}\right)$,
with weights $D_{t}\left(n\right),\,1\le n\le N$ chosen to emphasize
misclassified examples. The weight update rule is:\begin{equation}
D_{t+1}\left(n\right)=D_{t}\left(n\right)\left(q\left(t,+,X_{n}\right)e^{-\alpha_{t,+}y_{n}}+q\left(t,-,X_{n}\right)e^{\alpha_{t,-}y_{n}}\right)/Z_{t},\label{eq:Adaboost-update}\end{equation}
where $Z_{t}=\sum_{n=1}^{N}D_{t}\left(n\right)\left(q\left(t,+,X_{n}\right)e^{-\alpha_{t,+}y_{n}}+q\left(t,-,X_{n}\right)e^{\alpha_{t,-}y_{n}}\right)$
normalizes the weights so they sum to one.

With a deterministic weak classifier, one would have $q\left(\pm,X_{n}\right)\in\left\{ -1,+1\right\} $,
resulting in the original Adaboost weight update rule. An additional
difference is that the $q\left(\pm,X_{n}\right)$ are unknown. We
address this issue in Sec.~\ref{sub:Greedy-estimation-of-q} and
assume for now that the we have estimates $\hat{q}\left(\pm,X_{n}\right)$.

\subsection{Boosting property\label{sub:AdaboostBoosting}\vspace{-1mm}}

We now give an upper bound for the expected misclassification error
of $H\left(X\right)$, and show how to choose the weights $\alpha_{t,\pm}$.
This derivation parallels that of \cite{SchapireSinger99ML}: the
expected training error is \begin{equation}
E\left(\Loss\right)=\sum_{n=1}^{N}D\left(n\right)E\left(\dsb{H\left(X_{n}\right)\ne y_{n}}\right)\le\sum_{n=1}^{N}D\left(n\right)E\left(e^{-H\left(X_{n}\right)y_{n}}\right).\label{eq:ExponentialErrorBound}\end{equation}
where $\dsb{.}$ is the {}``indicator function,'' being 1 if the
bracketed expression is true and zero otherwise.

Since $H\left(X_{n}\right)$ may take at most $2^{T}$ possible values,
depending on the outputs $s_{t}$, $1\le t\le T$, of the $T$ classifiers
$h_{t}\left(X_{n}\right)$, one has:\begin{eqnarray*}
E\left(e^{-H\left(X_{n}\right)y_{n}}\right) & = & \sum_{s_{1},\ldots,s_{T}}\prod_{t=1}^{T}\hat{q}\left(t,s_{t},X_{n}\right)e^{-\alpha_{t,s_{t}}y_{n}}\\
 & = & \sum_{s_{1},\ldots,s_{T-1}}\prod_{t=1}^{T-1}\hat{q}\left(t,s_{t},X_{n}\right)e^{-\alpha_{t,s_{t}}y_{n}}\frac{D_{T+1}\left(n\right)Z_{T}}{D_{T}\left(n\right)}\\
 & = & \sum_{s_{1},\ldots,s_{T-2}}\prod_{t=1}^{T-2}\hat{q}\left(t,s_{t},X_{n}\right)e^{-\alpha_{t,s_{t}}y_{n}}\frac{D_{T}\left(n\right)Z_{T-1}}{D_{T-1}\left(n\right)}\frac{D_{T+1}\left(n\right)Z_{T}}{D_{T}\left(n\right)}\\
 & = & \ldots\textrm{etc}\ldots\\
 & = & \frac{D_{T+1}\left(n\right)}{D_{1}\left(n\right)}\prod_{t=1}^{T}Z_{t}.\end{eqnarray*}
Summing over all samples, one gets the familiar expression\[
E\left(\Loss\right)\le\prod_{t=1}^{T}Z_{t}.\]

The rest goes as with Adaboost: each $Z_{t}$ is minimized by setting
\[
\alpha_{t,+}=\frac{1}{2}\log\left(\frac{W_{t}^{++}}{W_{t}^{+-}}\right)\,\,\,\textrm{and}\,\,\,\alpha_{t,-}=\frac{1}{2}\log\left(\frac{W_{t}^{--}}{W_{t}^{-+}}\right),\]
where $W_{t}^{ab}=\sum_{n\mid y_{n}=b}D_{t}\left(n\right)\hat{q}\left(t,a,n\right)$,
for any $a\in\left\{ +1,-1\right\} ,\, b\in\left\{ +1,-1\right\} .$
For this choice of $\alpha_{t,\pm}$, one has $Z_{t}=2\sqrt{W_{t}^{++}W_{t}^{+-}}+2\sqrt{W_{t}^{-+}W_{t}^{--}}$,
and one can show that $Z_{t}\le\sqrt{1-4\eps^{2}}=\rho$. The expected
error of the $T$-stage boosted probabilistic classifier thus has
the same bound as the error of Adaboost:\begin{equation}
E\left(\Loss\right)\le\left(\sqrt{1-4\eps^{2}}\right)^{T}\stackrel{\Delta}{=}\rho^{T}\label{eq:AdaboostBound}\end{equation}

It must be made clear that, in practice, during training, the users
only have \emph{estimates} of $q\left(t,\pm,n\right)$, so that they
reduce \emph{an estimate} of the bound of the expected error.

\subsection{Estimation of $q\left(\pm,X_{n}\right)$ during training\label{sub:Greedy-estimation-of-q}\vspace{-1mm}}

In this section, we show how users can, in practice, balance their
need for accurate estimates of the $q\left(t,\pm,n\right)$ with their
eagerness to reduce the estimate on the bound of the expected error
(the reader may skip this part in a first reading).

The difference with respect to Adaboost is that, once the classifier
has been trained, the user has to estimate the $q\left(T,\pm,n\right)$,
in order to compute $D_{T+1}\left(n\right)$ for the next classifier.
The question is thus {}``how many samples of $h_{T}\left(X_{n}\right)$
should be taken?'' The trivial answer, which we exclude, is to fix
some number $R$ of samples and use the corresponding MAP or ML estimates.
We exclude for now this approach, to avoid adding an extra parameter
$R$. We propose, instead, two approaches based on the MAP estimator
of $q\left(t,\pm,n\right)$.

Let us first compare the MAP and ML estimators, to later better explain
our preference for the MAP. Both the MAP and ML converge in probability
to the true value, so that the corresponding estimators of $Z_{t}$
also converge in probability to the true value. The MAP and ML differ
in that the MAP estimator of $q\left(T,\pm,n\right)$ is biased towards
$\frac{1}{2}$, and that of $Z_{T}$ is biased towards $1$. More
precisely, the expected value of these MAP estimators converge to
the their limits \emph{from above}, so that the expected value of
successive estimates of $Z_{T}$ decreases towards the true value.
Thus, after sampling $h_{T}\left(X_{n}\right)$ R times, sampling
once more is always expected to decrease the estimate of $Z_{T}.$

\paragraph{First approach to estimate $q\left(T,\pm,n\right)$:}

The MAP thus has the advantage of providing a natural stopping time,
that of the first observed increase in our estimate of $Z_{T}$. The
event that $Z_{T}$ increases has a probability that increases towards
$1/2$, so that it will almost always (in the probabilistic sense)
happen after a finite time. This strategy can also be used with the
ML estimator, but, having a greater variance, it is more likely to
result in a spuriously low estimate of $Z_{T}$ and early stops. On
these grounds, the MAP should thus be preferred over the ML.

\paragraph{Second approach:}

An alternative method involving some look-ahead, and the user\'{ }s
stopwatch, may be also be considered: having until now trained $T$
classifiers and sampled $R$ times $h_{T}\left(X_{n}\right),\,1\le n\le N$,
the user has the following options:

\begin{description}
\item [A]Train a new classifier $h_{T+1}\left(X\right)$, using the current
estimate of $q\left(T,\pm,n\right)$ in the calculation of $D_{T+1}\left(n\right)$.
Then sample $h_{T+1}\left(X_{n}\right),\,1\le n\le N$ once, resulting
in a first MAP estimate of $Z_{T+1}$. As a result, the user decreases
the estimated bound, now $\prod_{t\le T+1}Z_{t}$, previously $\prod_{t\le T}Z_{t}$,
by the factor $Z_{T+1}$. Also, the user measured, with his or her
stopwatch, the elapsed time $S_{A}$ during training and sampling.
The instantaneous bound decrease rate per unit of time is $\left(Z_{T+1}\right)^{1/S_{A}}$.
\item [B]Sample once more $h_{T}\left(X_{n}\right),\,1\le n\le N$, producing
a new estimate $Z'_{T}$. As a result, the user decreases (or increases)
the estimated bound by a factor $Z'_{T}/Z_{T}$. Again, with his or
her stopwatch, (s)he measured the elapsed time $S_{B}$. The instantaneous
bound decrease rate is $\left(Z'_{T}/Z_{T}\right)^{1/S_{B}}$.
\end{description}
Finally, based on the smallest bound decrease rate, the user decides
whether to keep the new classifier $h_{T+1}\left(X\right)$ or the
new estimate $Z'_{T}$. 

We have thus proposed two parameterless ways to estimate $q\left(T,\pm,n\right)$.

\section{Boosting decision tree \label{sec:boosting-DT-model}\vspace{-1mm}}

Having shown how Adaboost can be transposed to probabilistic weak
learners, we now further extend our study to probabilistic decision
trees.

Computationally, our proposed classifier is a smoothed binary decision
tree. In that model, the output $H\left(X\right)$ is a weighed sum
of the (random) classifiers on the nodes traversed by an input element
$X$ :\begin{equation}
H\left(X\right)=\sum_{t=1}^{T\left(X\right)}\alpha_{s\left(t,X\right)}h_{s\left(t-1,X\right)}\left(X\right),\label{eq:ClassifModel}\end{equation}
where $s\left(t,X\right)$ is the index of the $t\nth$ node reached
by input $X$,~ $h_{s\left(t-1,X\right)}\left(X\right)\in\left\{ -1,1\right\} $
is the output of the corresponding classifier and $T\left(X\right)$
is the depth of the last inner node reached by $X$ before exiting
the decision tree. The weight given to $h_{s\left(t-1,X\right)}\left(X\right)$
, $\alpha_{s\left(t,X\right)}$ is domain-partitioned, since it depends
on the observed value of $h_{s\left(t-1,X\right)}\left(X\right)$.

Some notation is needed: the index of a node $s$ is a sequence of
{}``$+$'' and {}``$-$'', indicating the path to that node. For
example, in Fig.~\ref{fig:decisionTreeIndex}, right, $s=\left(+,+,-\right)$
is the leaf reached by following the {}``$+$'' edge out of the
root node, then the {}``$+$'' edge out of the $\left(+\right)$
node, then the {}``$-$'' edge out of the $\left(+,+\right)$ node.
The output of the classifier, when an input exits the decision tree
by $s$, is thus $H\left(X\right)=\alpha_{+}+\alpha_{++}+\alpha_{++-}$. 

For additional convenience, we write $\bar{s}$ the index of the parent
of node $s$ ($\bar{s}=\left(+,+\right)$ in the previous example)
and $\dot{s}$ the last edge followed to reach $s$ (here, $\dot{s}=-$).
Thus, one may write $s=\left(\bar{s},\dot{s}\right)$. The root node
is $s\left(0,X\right)=\emptyset$. With this notation, and noting
that $H\left(X\right)$ only depends on the leaf $l\left(X\right)$
reached by $X$, one has: \begin{equation}
H\left(X\right)=\sum_{\es<s\leq l\left(X\right)}\alpha_{s}\dot{s}\,\,\stackrel{\Delta}{=}\,\, H_{l},\label{eq:leaf-value}\end{equation}
where the sum is taken over all nodes $s$ between the leaf $l$ and
the root $\es$ (exclusive).

\subsection{Decision trees with probabilistic nodes\vspace{-1mm}\label{sub:Decision-trees-with}}

Like most other decision tree-building algorithms~\cite{Quinlan93C45,Mitchell97ML},
we add nodes one at a time and do not modify previously added nodes.
This is the most common way of avoiding the inherent complexity~\cite{HyafilRivest76IPL}
of building decision trees. We do not consider a subsequent pruning
step. Unlike other decision trees, and like in Adaboost, each node
is trained on the whole dataset. 

However, we modulate the weights of the examples, not only based on
whether they are misclassified (as in Adaboost), but also based on
their probability of reaching the node. After having trained $h_{s}\left(X\right)$
with weights $D_{s}\left(n\right)$, $1\le n\le N$, the weights for
training the children nodes $s+$ and $s-$ are:\begin{equation}
D_{s+}\left(n\right)=\frac{D_{s}\left(n\right)}{Z_{s+}}q\left(s+,X_{n}\right)e^{-\alpha_{s+}y_{n}}\,\,\textrm{and}\,\,\, D_{s-}\left(n\right)=\frac{D_{s}\left(n\right)}{Z_{s-}}q\left(s-,X_{n}\right)e^{-\alpha_{s-}y_{n}}.\label{eq:Adaboost-update}\end{equation}
In this expression, $Z_{sa}=\sum_{n=1}^{N}D_{s}\left(n\right)q\left(sa,X_{n}\right)e^{-\alpha_{sa}y_{n}}$,
for $a\in\left\{ +,-\right\} $, are normalizing constants, and $q\left(s+,X_{n}\right)\in\left[0,1\right]$
is the (unknown) parameter of the Bernoulli random variable $h_{s}\left(X\right)$.
Like in Sec.~\ref{sec:Adaboost}, we use estimates $\hat{q}\left(s+,X_{n}\right)$
in place of the true values.

\subsection{Bound on the expected error\vspace{-1mm}}

We now bound the error of the boosting tree algorithm and specify
the weights $\alpha_{s}$ and the choice of the trained node at each
step.

Using again the exponential error inequality $\Loss\left(H\left(X\right),y\right)\le e^{-H\left(X\right)y}$,
Eq.~(\ref{eq:ExponentialErrorBound}), the expected misclassification
error for a training example $X_{n}$ is upper-bounded by\begin{equation}
E\left(e^{-H\left(X_{n}\right)y}\right)=\sum_{l:\,\textrm{leaf}\, H}p\left(l,X_{n}\right)e^{-H_{l}y_{n}},\label{eq:treeIndividualError}\end{equation}
where $p\left(l,X\right)$ is the probability of an input $X$ reaching
the leaf $l$. More generally, assuming independence of the outputs
of classifiers at each node, the probability that $X$ reaches a node
$s=\left(s_{1},s_{2},\ldots,s_{T}\right)$ is \[
p\left(s,X\right)=q\left(s_{1},X\right)\cdot q\left(s_{1}s_{2},X\right)\cdot\ldots\cdot q\left(s_{1}\ldots s_{D},X\right)=\prod_{r\le s}q\left(r,X\right),\]
where the product is taken for all nodes between the root and $s$.

The error bound is thus\begin{eqnarray*}
E\left(e^{-H\left(X_{n}\right)y}\right) & = & \sum_{l:\,\textrm{leaf}\, H}\left(\prod_{s\le l}q\left(s,X_{n}\right)\right)e^{-H_{l}y_{n}}\\
 & = & \sum_{l:\,\textrm{leaf}\, H}\prod_{s\le l}\left(q\left(s,X_{n}\right)e^{-\dot{s}\alpha_{s}y_{n}}\right)\\
 & = & \sum_{l:\,\textrm{leaf}\, H}\prod_{s\le l}\left(\frac{D_{s}\left(n\right)}{D_{\bar{s}}\left(n\right)}Z_{s}\right)\\
 & = & \sum_{l:\,\textrm{leaf}\, H}D_{l}\left(n\right)\prod_{s\le l}Z_{s}\end{eqnarray*}
Summing over all examples $X_{n}$, $1\le n\le N$ and replacing in
Eq.~(\ref{eq:ExponentialErrorBound}) yields the bound:\begin{equation}
E\left(\Loss\right)\le\sum_{l:\,\textrm{leaf}\, H}\prod_{s\le l}Z_{s}\label{eq:AdatreeBound}\end{equation}

Like above, each $Z_{s}$ is minimized by setting \[
\alpha_{s}=\frac{1}{2}\log\left(\frac{W_{\bar{s}}^{\dot{s}\dot{s}}}{W_{\bar{s}}^{\dot{s}\neg\dot{s}}}\right),\,\,\,\textrm{where}\,\,\, W_{\bar{s}}^{ab}=\sum_{n\mid y_{n}=b}D_{\bar{s}}\left(n\right)q\left(a,n\right),\,\, a,b\in\left\{ +,-\right\} ,\]
and $\neg$ is the negation operator. For these values of $\alpha_{s}$,
each $Z_{s}$ is takes the value \[
Z_{s}=2\sqrt{W_{\bar{s}}^{\dot{s}+}W_{\bar{s}}^{\dot{s}-}}.\]

This bound can also be found, in slightly different contexts, in our
previous work~\cite{Grossmann04LCVPR} and in our unpublished manuscript~\cite{Grossmann04Adatree2}.
In the present paper, we additionally study how this bound evolves
with the size of the tree.

\subsubsection*{Expected error bound as a function of the tree size}

We now describe the evolution of the bound (\ref{eq:AdatreeBound})
when the tree is grown by a greedy bound-reducing algorithm.

As previously, we may show that $Z_{s+}+Z_{s-}\le\sqrt{1-4\eps^{2}}=\rho$,
owing to the probabilistic weak learner hypothesis.

We now proceed recursively. After training and incorporating $T$
nodes, the expected error bound is $C\left(T\right)=\sum_{l:\,\textrm{leaf}\, H}\prod_{s\le l}Z_{s}$.
At this point, the tree has $T+1$ leaves, so that one leaf $l$ at
least has an error not less than $C\left(T\right)/\left(T+1\right)$.
After training a probabilistic weak classifier at $l$, the new error
bound is \begin{eqnarray*}
C\left(T+1\right) & = & C\left(T\right)-\prod_{s\le l}Z_{s}+\prod_{s\le l-}Z_{s}+\prod_{s\le l-}Z_{s}\\
 & = & C\left(T\right)+\left(\prod_{s\le l}Z_{s}\right)\left(-1+Z_{l+}+Z_{l-}\right)\\
 & \le & C\left(T\right)+\left(\frac{C\left(T\right)}{T+1}\right)\left(-1+\rho\right)\\
 & = & C\left(T\right)\left(\frac{T+\rho}{T+1}\right).\end{eqnarray*}
Since $C\left(0\right)=1$, we have the general relation\begin{equation}
C\left(T\right)\le\prod_{t=0}^{T}\frac{t+\rho}{t+1}=\frac{1}{TB\left(T,\rho\right)}\stackrel{\Delta}{=}F\left(T,\rho\right)\simeq\frac{T^{\rho-1}}{\Gamma\left(\rho\right)},\label{eq:AdatreeBadBound}\end{equation}
where $B\left(T,\rho\right)$ is the beta function and $\Gamma\left(\rho\right)$
is the Gamma function. The rightmost term is the asymptotic approximation
for large $T$; it is coherent with the bound of~d\cite[Eq. 6]{KearnsMansour99JCSS}.

\begin{figure}
\begin{spacing}{0}
\begin{center}\includegraphics[%
  width=0.5\textwidth]{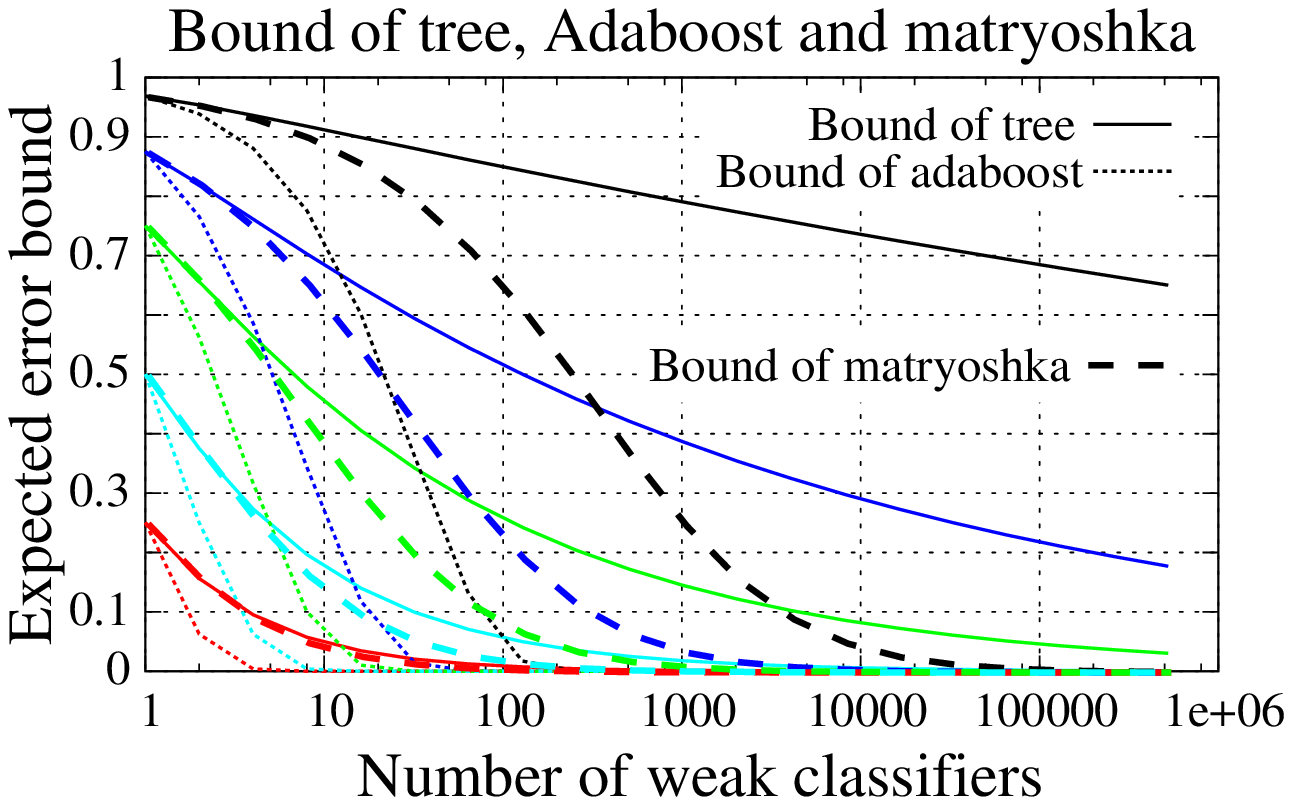}~~~\includegraphics[%
  width=0.5\textwidth,
  height=1.5in]{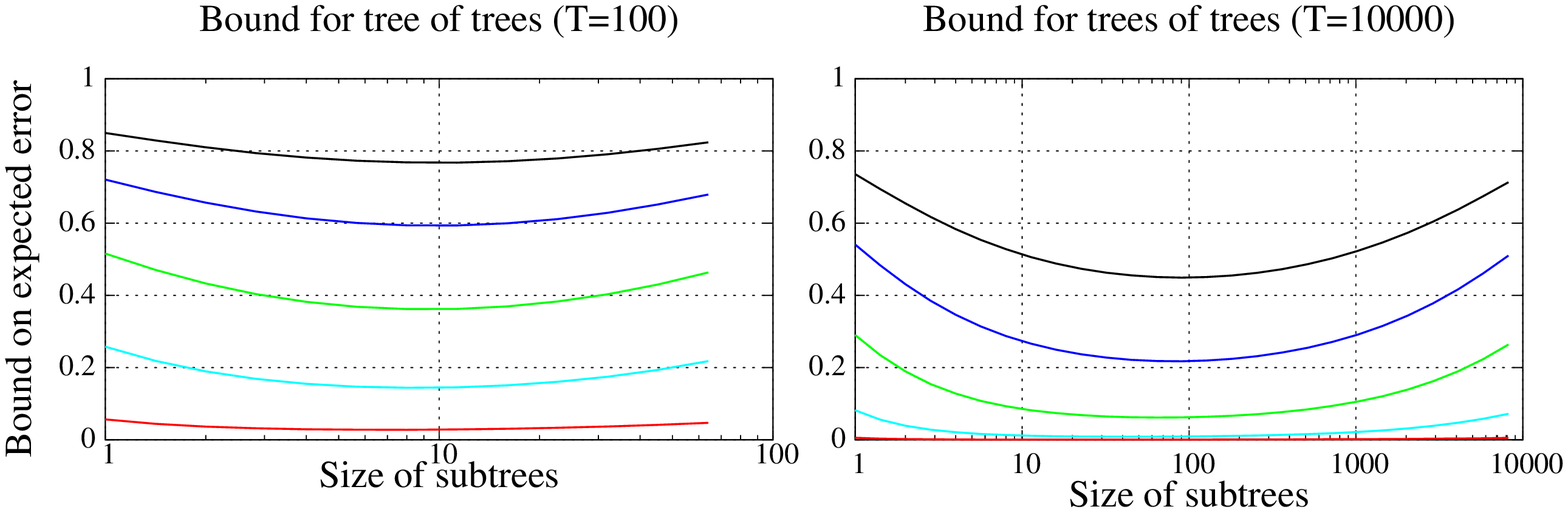}\vspace{-3mm}\end{center}
\end{spacing}

\caption{Left: Bound of boosted decision tree (full curve, highest), Eq.~(\ref{eq:AdatreeBadBound}),
of probabilistic Adaboost (dotted, lowest), Eq.~(\ref{eq:AdaboostBound}),
and of matryoshka (dashed, middle). From top to bottom, $\rho\in\left\{ \frac{31}{32}\right.,$$\frac{7}{8},$$\frac{3}{4},$$\frac{1}{2},$$\left.\frac{1}{4}\right\} $,
i.e. $\eps\in\left\{ 0.12\right.,$$0.24,$$0.33,$$0.43$$,\left.0.46\right\} $.
Right: Bound of boosted tree of simple trees, given by Eq.~(\ref{eq:BoundTree2}).
\vspace{-3mm}\label{cap:Bound-AT-AB-Matryoshka}\label{cap:Bound-Tree2}\label{cap:Bound-AT-AB}}
\end{figure}

This bound is interesting in more than one respect:

\begin{itemize}
\item It appears that it cannot be very much improved, in the following
sense: consider a probabilistic learner with error $1/2-\eps$, independently
of the weights $D\left(n\right)$ with which it is trained. This learner
verifies the probabilistic weak learner hypothesis. Now, for both
the probabilistic Adaboost and for a (balanced) decision tree, $H\left(X\right)$
is a binomial random variable with parameters $\left(1/2-\eps\right)$,
and the number of weak parameters traversed by $X$. This second parameter
is $T$ for Adaboost and $\log_{2}\left(T\right)$ for the decision
tree. It is clear, then, that the decision tree requires exponentially
more weak classifiers than the probabilistic Adaboost.
\item This bound is especially bad for very weak classifiers ($\rho\simeq1$).
The full curves in Figure~\ref{cap:Bound-AT-AB}, left, plot the
bound $F\left(T,\rho\right)$ for $\rho=31/32$, 7/8, 3/4, 1/2 and
1/4. For comparison, the expected error bound of Adaboost, $\rho^{T}$,
plotted alongside, is much lower, especially for $\rho=31/32$.
\item This bound calls the attention of designers of decision trees tempted
to pass all the training dataset along all branches: if the weak classifier
is very weak, the number of needed weak classifiers may grow very
much. With stronger classifiers, the boosting tree algorithm may be
more practical.
\end{itemize}

\section{Matryoshka decision trees\label{sec:Matryoshka}\vspace{-1mm}}

Based on the conclusion of the previous section -that stronger classifiers
yield better boosted decision trees, we now address the question of
obtaining sufficiently strong classifiers. The first step in this
direction (Section~\ref{sub:Bound-Tree2}) is to explore the idea
of putting a boosted tree at each node. We will see that there is
an advantage in doing so. It will then be natural, in Section~\ref{sub:Bound-Matryoshka},
to build trees of trees of trees of ... of weak classifiers, that
is, a matryoshka of decision trees.

\subsection{Bound for a tree of trees\label{sub:Bound-Tree2}\vspace{-1mm}}

In this section, we study the error bounds obtainable by a decision
tree built using the method of Section~\ref{sec:boosting-DT-model},
but where the nodes are themselves trees built according to that same
method. We place ourselves in the situation of having the resource
to train a fixed number $T$ of weak classifiers, and our objective
is to minimize the bound on the expected error.

In this context, it is natural to study the bound obtainable by assembling
$T_{2}$ sub-trees of fixed size $T_{1}=T/T_{2}$. By Eq.~(\ref{eq:AdatreeBadBound}),
the error bound for the sub-trees is $F\left(T_{1},\rho\right)=\left(T_{1}B\left(T_{1},\rho\right)\right)^{-1}$,
and that of the outer tree is \begin{equation}
F\left(\frac{T}{T_{1}},F\left(T_{1},\rho\right)\right).\label{eq:BoundTree2}\end{equation}
Figure~\ref{cap:Bound-Tree2}, right, plots this bound plotted against
$T_{1}$. The curves show that, for $T_{1}=1$ and $T_{1}=T$, the
bound is the same as $F\left(T,\rho\right)$, i.e. that of a not-nested
decision tree. More interestingly, for intermediate values of $T_{1}$,
the bound of Eq.~(\ref{eq:BoundTree2}) is always lower than $F\left(T,\rho\right)$.
In particular, the minimum is always near $T_{1}=\sqrt{T}$. 

Given these encouraging results, we are naturally tempted to substitute
the sub-trees (of size $T_{1}$) by$T_{1}'$ sub-trees of sub-sub-trees
of size $T''_{1}$, for some $T_{1}'$, $T_{1}''$ s.t. $T_{1}'T_{1}''=T_{1}$.
The same idea can also be applied to the outer tree.

\subsection{Bound for a tree of trees ... of trees of weak classifiers\vspace{-1mm}\label{sub:Bound-Matryoshka}}

More generally, we are tempted to determine the bounds reachable by
trees of trees of ... of trees of weak classifiers. For some $L$
and $T_{1},$$T_{2}$,$\ldots$$T_{L}$ s.t. $T_{1}T_{2}\ldots T_{L}=T$,
the bound is easily shown to be:\[
F\left(T_{L},F\left(T_{L-1},\ldots F\left(T_{1},\rho\right)\right)\right).\]
Finding analytically the optimal combination of $T_{i}$, $1\le i\le L$
may not be easy. But, guided by the observation that, for $L=2$,
the optimal choice seems to be near $T_{1}=T_{2}=\sqrt{T}$, we naturally
consider the case $T_{1}=T_{2}=\ldots=T_{L}=T^{1/L}$. In this case,
the bound is\begin{equation}
F\left(T^{\frac{1}{L}},F\left(T^{\frac{1}{L}},\ldots F\left(T^{\frac{1}{L}},\rho\right)\right)\right).\label{eq:Bound-Iso-Sized}\end{equation}

\begin{figure}
\begin{center}\includegraphics[%
  width=0.4\textwidth]{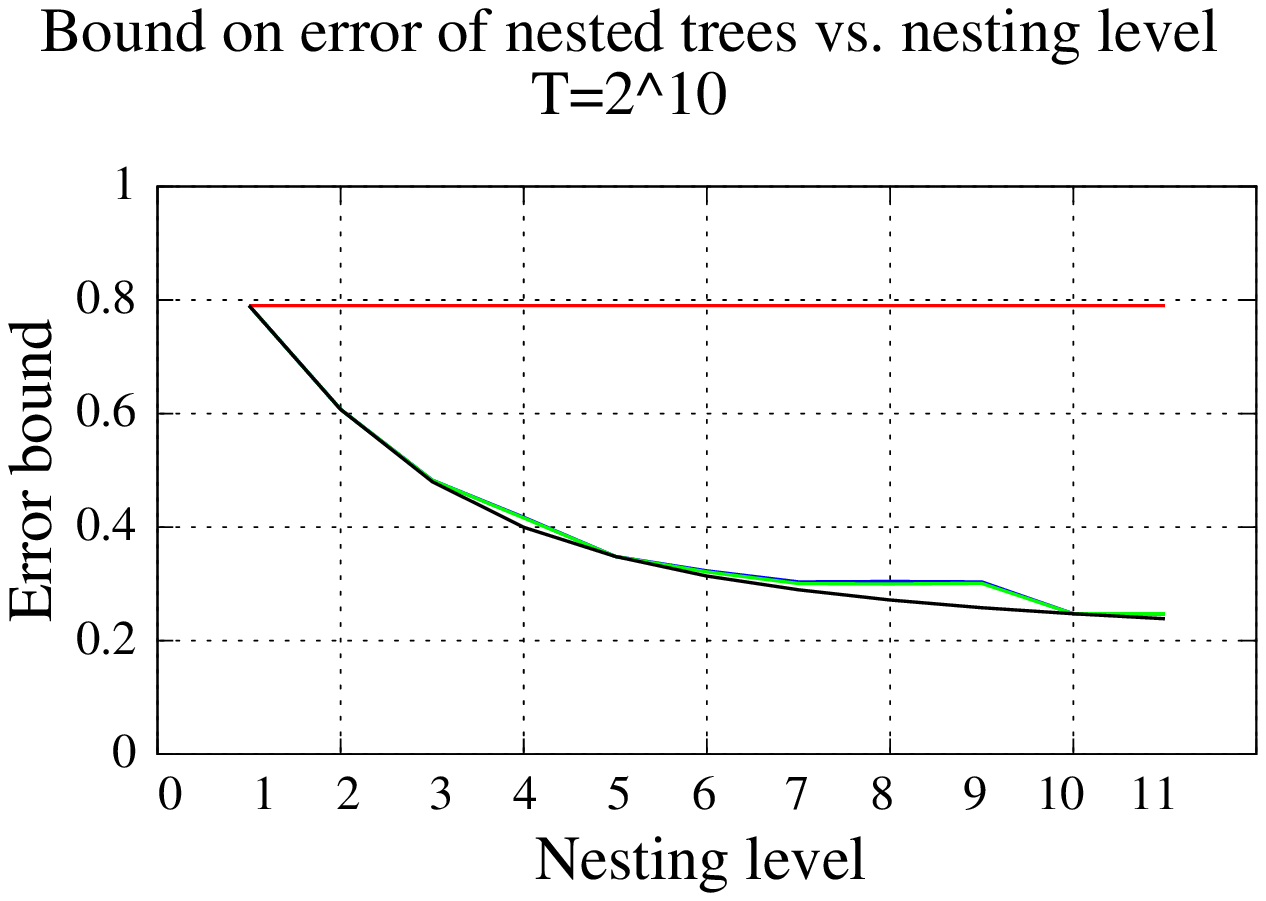}\includegraphics[%
  width=0.4\textwidth]{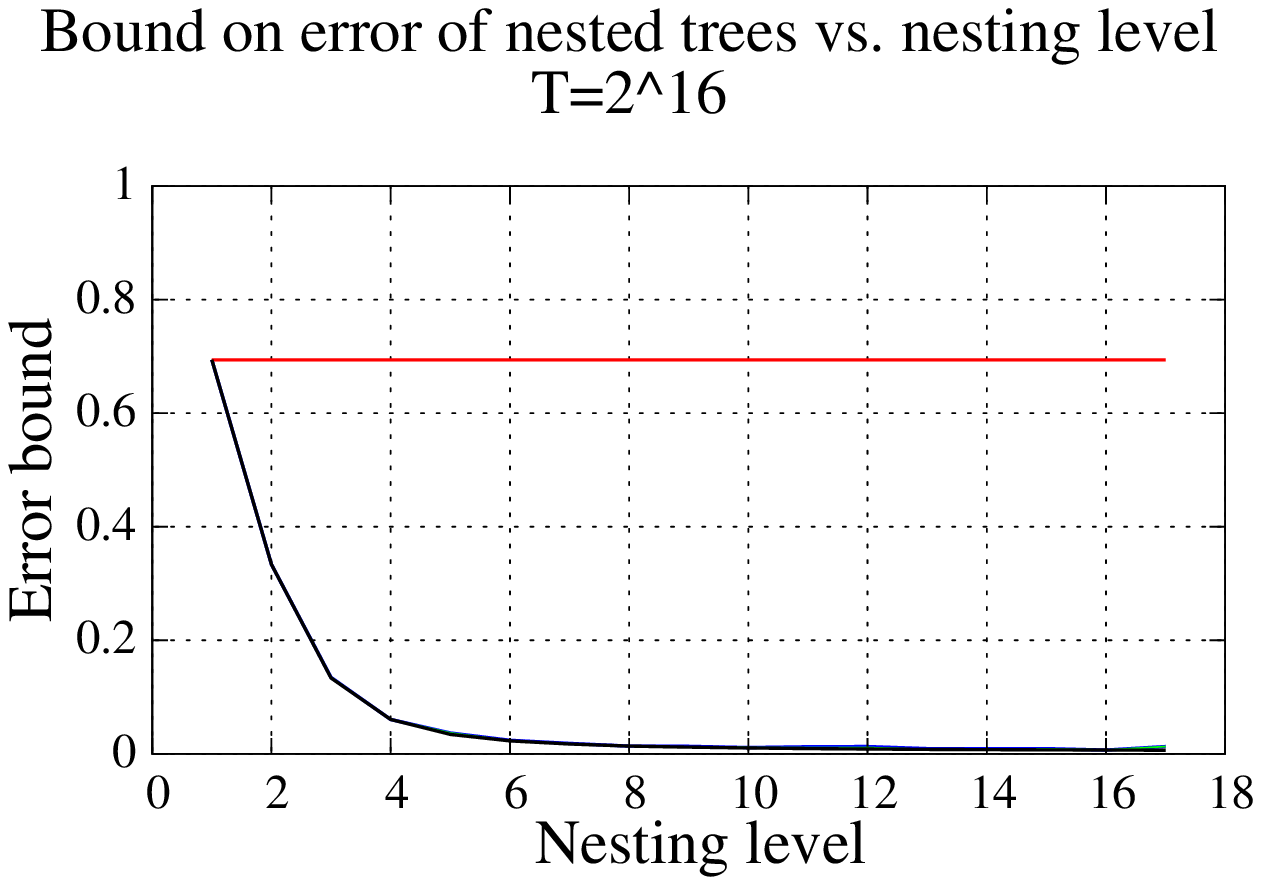}\vspace{-5mm}\end{center}

\caption{Black curve: bound on error of matryoshka decision trees at various
levels of nesting. The sub-tree sizes are $T^{1/\left(\textrm{nesting level}\right)}$.
Light-colored curves near the black curve are for trees w/ integer
number of nodes. . The topmost line marks the error bound of the (not
nested) decision tree. At left, $T=1024$, and $T=65536$ at right.
These curves are for $\rho=\frac{31}{32}$ i.e. $\eps\simeq0.12$.\label{cap:Bound-Nested-Level}\vspace{-3mm}}
\end{figure}

The black graph in Figure~\ref{cap:Bound-Nested-Level} plots this
value against the nesting level $L$, with the original bound $F\left(T,\rho\right)$
(top) for comparison. This figure clearly shows that deeper nesting
levels improve the bound. In fact, Eq.~(\ref{eq:Bound-Iso-Sized})
continues to decrease for $L>\log_{2}T$, i.e. when the trees each
have less than two nodes.

This (strange) effect is due to the fact that $F\left(T,\rho\right)$
is defined for any positive real $T$. Since the number of nodes is
in an integer, there are no practical repercussions.

However, these curves clearly indicate that smaller sub-trees yield
better bounds. This suggests building the smallest possible trees,
with just two nodes, each node a tree with two nodes, etc, until the
last level, consisting of trees with two weak classifiers.

\subsection{Bound for 2-matryoshka\vspace{-1mm}}

We now derive the expected error bound for the {}``2-matryoshka''
tree, having exactly two nodes, at all nesting levels, having precisely
two nodes. We thus need to assume that $T=2^{L}$ is a power of two.

We call $M_{2}\left(T,\rho\right)=F\left(2,F\left(2,\ldots F\left(2,\rho\right)\right)\right)$
the bound for this tree (there are $L$ nested parentheses). Recalling
from Eq.~(\ref{eq:AdatreeBadBound}) that $F\left(2,\rho\right)=\rho\frac{1+\rho}{2}=\frac{1}{2}\rho+\frac{1}{2}\rho^{2}$,
one writes $M_{2}\left(T,\rho\right)$ as a polynomial of degree $T$.

Figure~\ref{cap:Bound-AT-AB-Matryoshka}, left, shows the graph of
$M_{2}\left(T,\rho\right)$, in dashed lines. This figure shows that
the 2-matryoshka tree has a much stronger boosting ability than the
plain boosting tree, and this is the main result of this paper.\nopagebreak

\subsection{Building a matryoshka\vspace{-1mm}}

The algorithm for the 2-matryoshka would thus be: train a two-leaf
tree (stage {}``b'', in Fig.~\ref{cap:Nested-decision-tree}),
and collect the leaves into a single node ({}``c''). Train a two-leaf
sub-tree on one of the branches, collect its leaves in a single node
({}``e''). Collect the leaves once more ({}``f'') etc. If all
weak classifiers have the same edge $\eps$, then this approach is
the most appropriate.

In practice, the classifiers will not have the same edge and a greedy
-with respect to number of nodes or physical training time- bound-decreasing
approach could be considered. Each time a classifier is added to the
tree, we will consider each sub-tree containing that node, starting
from the top. For each sub-tree, we compare the instantaneous bound
decrease rate%
\footnote{Here, we consider the decrease rate per added node, but the decrease
rate per unit of training time could be used too.%
} of the sub-tree at $T$, \begin{equation}
\dot{C}_{Simple}\simeq\left(C\left(T+1\right)-C\left(T-1\right)\right)/2,\label{eq:SimpleBoundDecRate}\end{equation}
 (\mycomment{$T+1$ being the number of nodes in the sub-tree, }$C\left(T\right)$
being computed on the sub-tree only), with that of a tree having such
a sub-tree at each node, \begin{equation}
\dot{C}_{Matryoshka}=\frac{\partial}{\partial t}F\left(\frac{t}{T},C\left(T\right)\right)\left(t=T\right).\label{eq:MatryoshkaBoundDecRate}\end{equation}
 If the later is smaller, then the leaves of the sub-tree are collected
into a single node.

We now give the detail of computing Eq.~(\ref{eq:MatryoshkaBoundDecRate}).
Using the relation $\frac{\partial}{\partial x}B\left(x,y\right)=B\left(x,y\right)\left(\psi\left(x\right)-\psi\left(x+y\right)\right)$,
where $\psi$ is the digamma function, $\psi\left(x\right)=\frac{\partial}{\partial x}\left(\log\left(\Gamma\left(x\right)\right)\right)$,
one gets\begin{eqnarray*}
F'_{T}\left(T,\rho\right) & = & -F\left(T,\rho\right)\left(\frac{1}{T}+\psi\left(T\right)-\psi\left(T+\rho\right)\right)\,\textrm{and}\\
F'_{\rho}\left(T,\rho\right) & = & -F\left(T,\rho\right)\left(\psi\left(\rho\right)-\psi\left(T+\rho\right)\right).\end{eqnarray*}
The first line above then gives

\begin{equation}
\frac{\partial}{\partial t}F\left(\frac{t}{T},C\left(T\right)\right)\left(t=T\right)=\frac{C\left(T\right)}{T}\left(\gamma+\psi\left(C\left(T\right)\right)+\frac{1}{C\left(T\right)}-1\right),\label{eq:DecreaseRateMatryoshka}\end{equation}
where $\gamma=-\psi\left(1\right)\simeq0.5772$ is Euler\'{ }s constant.

One can check that, for $T=1$, $\dot{C}_{Simple}=\dot{C}_{Matryoshka}$
and that, if $C\left(T\right)=F\left(T,\rho\right)$, i.e. if the
bound~(\ref{eq:AdatreeBadBound}) is tight, then $\dot{C}_{Simple}=\dot{C}_{Matryoshka}$
for all $T>1$.

\section{Discussion and conclusions\label{sec:Conclusions}\vspace{-1mm}}

We have developed in this paper a theory of probabilistic boosting,
aimed at decision trees. We proposed a boosting tree algorithm and
a theoretically superior matryoshka decision tree algorithm. These
algorithms are essentially parameter-free, owing to the principle
of choosing whichever training action most reduces the expected training
error bound, and to a judicious choice of possible training actions.

We showed bounds on the expected training error of the algorithms,
one of them discouraging, the other, encouraging. The bounds for simple
trees and for trees of trees are coherent with our early experiments.

Future developments include an analysis of the effect of approximating
the node branching probabilities $q\left(s,X_{n}\right)$ during training
and experimental evaluation of the matryoshka.

On a more general level, we believe that the high bound for boosting
trees indicates that the probabilistic weak learner hypothesis is
inadequate. This hypothesis, directly adapted from the theory of boosting,
does not take into account the fact that real-world classifiers usually
have a lower training error on smaller training sets. Our intuition
is thus that the entropy of the training weights, $D\left(n\right)$,
should be taken into account in future work.\vspace{-1mm}

\small 

\bibliographystyle{unsrt}
\bibliography{/home/etienne/paper/refs/allmybib}


\end{document}